\pdfoutput=1
\documentclass[11pt]{article}

\usepackage[]{acl}

\usepackage{times}
\usepackage{latexsym}
\usepackage{tabularx}
\usepackage{booktabs}
\usepackage{algorithm}
\usepackage{hyperref}
\usepackage{multicol}
\usepackage{fancyvrb}
\usepackage{tcolorbox}
\tcbuselibrary{breakable}
\usepackage[T1]{fontenc}
\usepackage{xcolor}
\usepackage{soul}
\usepackage[utf8]{inputenc}

\usepackage{algorithm}
\usepackage{algpseudocode}
\usepackage{microtype}

\usepackage{graphicx}
\usepackage{xcolor} % for colors
\usepackage{todonotes}
\usepackage{graphicx}
\usepackage{mwe} % provides example images (including the duck)
\usepackage{booktabs}
\usepackage{subcaption}  % for subtable environments
\usepackage{tabularx}    % for wrapping long rationale text
\usepackage{array} 
\usepackage{multirow}
\usepackage{makecell}
\usepackage{amsfonts}
\usepackage{amsmath}
\newcommand{\rescell}[2]{\makecell{$#1{\scriptstyle \pm#2}$}}
\usepackage{booktabs}

\usepackage{adjustbox}

\newcommand{\DPOSFTaug}{\(DPO_{rub}^{\pi_{sft}}\)}
\newcommand{\SFT}{SFT}
\newcommand{\DPOSFTgt}{\(DPO_{gt}^{\pi_{sft}}\)}
\newcommand{\fone}{\(F_1(\hat{\rho}_{\pi})\)}

\usepackage{enumitem}

\newlist{rublist}{itemize}{1}
\setlist[rublist]{left=0pt, label=--, labelsep=0.6em, itemsep=0.2ex, topsep=0.2ex, align=left}

\usepackage{inconsolata}
\title{The Subjectivity of Respect in Police Traffic Stops: \\ Modeling Community Perspectives in Body-Worn Camera Footage}

\author{
\textbf{Preni Golazizian}$^{1}$,
\textbf{Elnaz Rahmati}$^{1}$, 
\textbf{Jackson Trager}$^{2}$,
\textbf{Zhivar Sourati}$^{1}$, \\
\textbf{Nona Ghazizadeh}$^{1,2}$,
\textbf{Georgios Chochlakis}$^{1}$, 
\textbf{Jose Alcocer}$^{6}$\thanks{These authors contributed equally to this work.},
\textbf{Kerby Bennett}$^{4,5*}$,
\textbf{Aarya Vijay Devnani}$^{1*}$, \\
\textbf{Parsa Hejabi}$^{1*}$, 
\textbf{Harry G. Muttram}$^{3*}$,
\textbf{Akshay Kiran Padte}$^{1*}$,
\textbf{Mehrshad Saadatinia}$^{1*}$,
\textbf{Chenhao Wu}$^{1*}$, \\
\textbf{Alireza S.\ Ziabari}$^{1*}$, 
\textbf{Michael Sierra{-}Arévalo}$^{7}$\thanks{Senior authors.},
\textbf{Nick Weller}$^{3\dagger}$,
\textbf{Shrikanth Narayanan}$^{1,2\dagger}$, \\
\textbf{Benjamin A.\ T.\ Graham}$^{5\dagger}$ \and
\textbf{Morteza Dehghani}$^{1,2\dagger}$ 
\\
\textsuperscript{1}Department of Computer Science, University of Southern California, \\
\textsuperscript{2}Department of Psychology, University of Southern California, \\
\textsuperscript{3}Department of Political Science, University of California Riverside, \\
\textsuperscript{4}Department of Anthropology, University of California Los Angeles, \\
\textsuperscript{5}Department of Political Science and International Relations, University of Southern California, \\
\textsuperscript{6}Harvard Law School, Harvard University,
\textsuperscript{7}Department of Sociology, The University of Texas at Austin
\\
\texttt{golazizi@usc.edu} \\
}

\begin{document}

\maketitle
% \addtolength{\topskip}{4em}
\begin{abstract}

Traffic stops are among the most frequent police–civilian interactions, and body-worn cameras (BWCs) provide a unique record of how these encounters unfold. Respect is a central dimension of these interactions, shaping public trust and perceived legitimacy, yet its interpretation is inherently subjective and shaped by lived experience, rendering community-specific perspectives a critical consideration. 
Leveraging unprecedented access to Los Angeles Police Department BWC footage, we introduce the first large-scale traffic-stop dataset annotated with \emph{respect ratings} and free-text \emph{rationales} from multiple perspectives. By sampling annotators from police-affiliated, justice-system-impacted, and non-affiliated Los Angeles residents, we enable the systematic study of perceptual differences across diverse communities.
To this end, \textit{i)} we develop a domain-specific evaluation rubric grounded in procedural justice theory, LAPD training materials, and extensive fieldwork; \textit{ii)} we introduce a rubric-driven preference data construction framework for perspective-consistent alignment, and \textit{iii)} we propose a perspective-aware modeling framework that predicts personalized respect ratings and generates annotator-specific rationales for both \emph{officers} and \emph{civilian drivers} from traffic-stop transcripts.
Across all three annotator groups, our approach improves both rating prediction performance and rationale alignment. Our perspective-aware framework enables law enforcement to better understand diverse community expectations, providing a vital tool for building public trust and procedural legitimacy.

\end{abstract}

\section{Introduction}

\begin{figure*}[t]
  \centering
  \includegraphics[width=\linewidth]{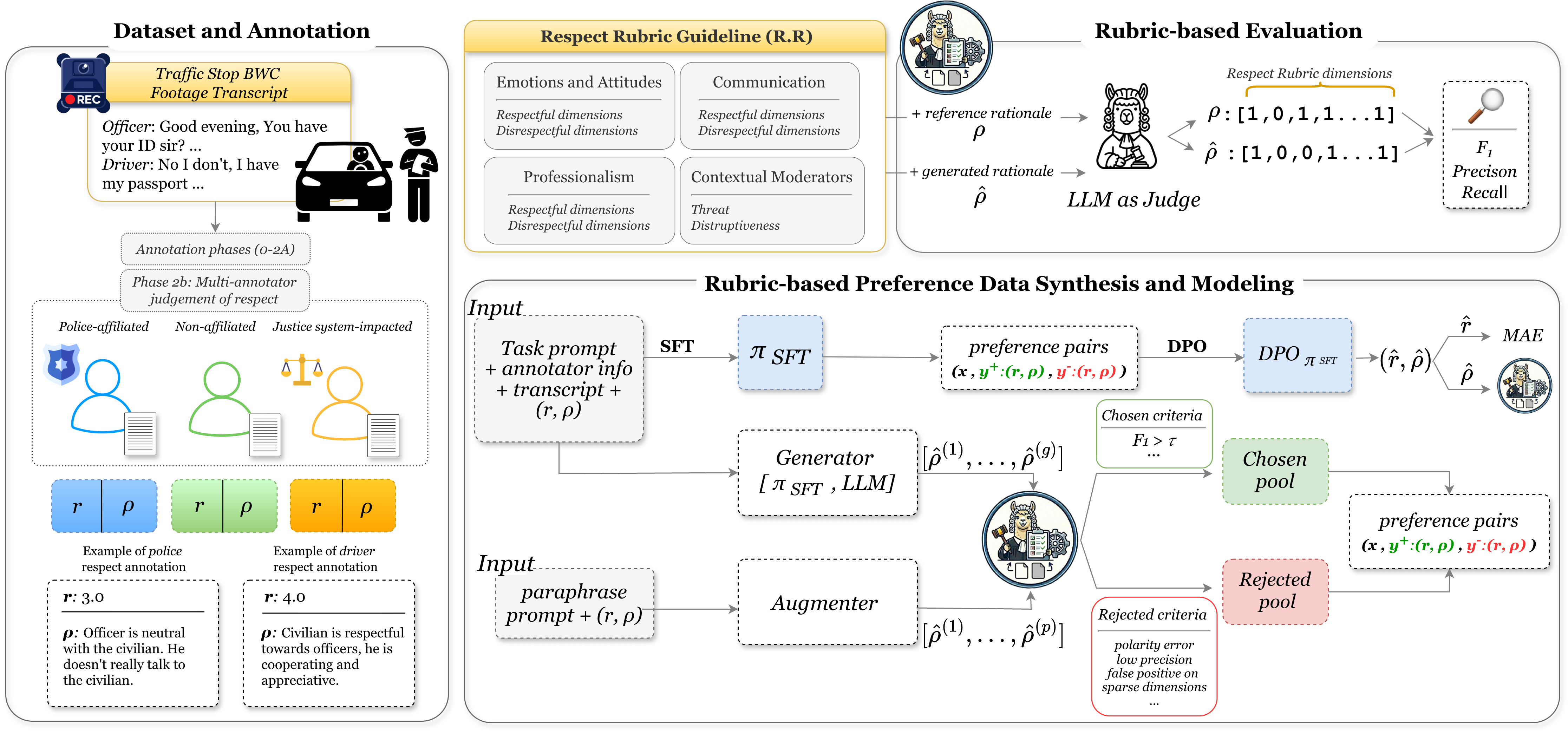} 
  \caption{Overview of the proposed framework, illustrating multi-perspective respect annotation from BWC transcripts, and domain-specific rubric development. The framework integrates rubric-guided preference data synthesis with supervised fine-tuning and alignment to produce perspective-aware respect ratings and rationales.}
  \label{fig:main}
  % \vspace{-0.5cm}
\end{figure*}

Police traffic stops are common, but fraught interactions between the public and their government. When officer-driver communication is ineffective, it undermines public trust and may lead to violent outcomes \citep{tyler2002trust}. At present, the vast majority of stops in major police jurisdictions in the U.S. are captured on bodyworn cameras (BWCs), a practice now becoming common worldwide
\citep[e.g.,][]{laming2019police}, offering high-quality data to understand and improve these interactions. The sheer volume of data, however, precludes the possibility of manual review. 

The interpretation of complex social interactions, such as police-civilian encounters, is inherently \emph{subjective}. 
 Perspectives on good communication vary across communities, influenced by lived experiences and systemic differences rooted in past encounters with the justice system 
\citep{sierra2025police}.
% \citep{anonymous2024}.
For example, previous research shows that Black people experience more frequent and more negative police interactions \citep{pierson2020large,xu2024racial}, affecting perceptions of police legitimacy and potentially contributing to fear, anxiety, and distrust within communities %\citep{weaver2010political,pickett2022american}.
\citep{pickett2022american}.

Prior research has established \emph{respect} as a measurable and consequential dimension of police-civilian interactions \citep[e.g.,][]{voigt2017language,camp2024body}. What remains less explored is how differences in lived experience shape interpretation and how to incorporate such variation into computational models.
This gap is particularly critical because when the subjectivity of core concepts, like respect, is ignored in ML, efforts to establish a single ground truth have historically privileged the perspectives of the dominant institutional and majority-group perspectives  \citep{turner2018detecting,barabas2020studying}. This focus on a single perspective overlooks how respect is interpreted differently across communities and limits the ability of models to reflect these differences.

In this work, we focus on \emph{respect}, as expressed by both officers and drivers. Respect is a central dimension of police-civilian interactions, influencing public trust and the perceived legitimacy of the state's authority %\citep{tyler2002trust, peffley2010justice, voigt2017language}, 
\citep{tyler2002trust, voigt2017language},
and consistently emerging as a key factor in how individuals interpret encounter outcomes. This focus is built upon established procedural justice literature, extensive survey, and qualitative interviews in Los Angeles \citep{sierra2025police}.
% \cite{anonymous2024}.
% This focus is built upon established procedural justice literature and extensive survey and qualitative interviews in Los Angeles 

Recognizing the subjective nature of respect in police-driver interactions, it is essential to employ subjective computational modeling techniques that accommodate multiple plausible perspectives and interpretations. Therefore, %treating labeling disagreements as meaningful signals %\citep{chochlakis2025humans, anonymous2024}, 
we undertook an 18-month effort to annotate Los Angeles Police Department (LAPD) BWC data from a demographically and experientially diverse pool of annotators: %\citep{anonymous2024}: 
 \emph{police-affiliated} individuals, \emph{justice-system-impacted} individuals, and other Los Angeles residents. This multi-perspective approach captures both shared and differing views of respect. %across communities. %The resulting dataset includes numeric ratings and free-text rationales, providing a unique resource for studying respect in everyday policing encounters.

In this work: \textit{(1)} we introduce a large-scale dataset of annotations of LAPD traffic-stop BWC footage, consisting of multi-annotator, fine-grained respect ratings and rationales collected from a demographically and experientially diverse pool of annotators. The dataset reflects the subjective and heterogeneous ways community members interpret officer and civilian behavior. 
\textit{(2)} we develop a domain-specific, theory- and data-grounded \emph{respect rubric} that synthesizes community perspectives, LAPD training materials, and prior research to characterize respectful and disrespectful communicative behaviors in traffic stops. The rubric is operationalized through an LLM-as-a-judge framework to evaluate human and model-generated rationales, enabling interpretable comparison and the construction of a rubric-guided preference dataset.
\textit{(3)} we model the subjective, group- and individual-dependent nature of perceived respect with a single annotator-aware and group-aware model, which conditions on annotator background to personalize predictions across %\textit{police-affiliated}, \textit{justice-impacted}, and \textit{non-impacted/affiliated} 
community members.

Our framework, as illustrated in Figure~\ref{fig:main}, predicts respect ratings and annotator-specific rationales that reflect group’s interpretive lens, with rubric-guided preference data improving model alignment and rationale quality across groups.

\section{The LAPD Respect Dataset} \label{sec:data}

Our dataset comprises BWC videos and their annotations from a stratified random sample of about 1,000 LAPD traffic stops recorded between 09/2021- 09/2022. These videos and their annotations were collected through a large-scale, multi-year collaboration with LAPD, facilitated by the Office of the Inspector General. Footage is recorded from officers' BWCs, which are typically activated at the start of a stop, though coverage can vary due to activation timing and technical factors.

We collect fine-grained annotations from a diverse pool of trained annotators (§\ref{sec:annotator_setting}) across multiple phases (§\ref{sec:phases}). This work focuses on the phase involving \emph{respect} ratings and free-text \emph{rationales} for both officers and civilians, using conversation transcripts as the primary input modality.
 \subsection{Annotator Selection and Setting} 
\label{sec:annotator_setting}
The core objective of this project is to systematically document and model the heterogeneity of community perceptions regarding police-civilian encounters. To achieve this, we have conducted extensive fieldwork, including officer and community focus groups and interviews, a representative survey of over 2,000 LA residents, an analysis of LAPD training manuals, and ride-alongs with LAPD officers. %\citep{anonymous2024}.
% \citep{sierra2025police}.
Crucially, these diverse engagements reveal that individuals hold different, and often conflicting, preferences for the same communicative behaviors; this finding confirms previous research 
%\citep[e.g.,][]{tyler2004enhancing,weaver2010political} 
\citep[e.g.,][]{weaver2010political} that lived experiences with the justice system strongly shape the interpretation of respect, validating the central premise of our subjective modeling approach.
Based on our fieldwork, annotators are recruited from three groups:
% \begin{itemize}[leftmargin=0.5em,noitemsep, topsep=1pt]
%     \item \textit{Police-affiliated} (\(G_{PA}\)): individuals with prior law enforcement experience,
%     \item \textit{Justice system–impacted} (\(G_{JI}\)): individuals with a history of incarceration or arrest,
%     \item \textit{Non-affiliated} (\(G_{NA}\)): community members without prior law enforcement experience or history of incarceration/arrest.
% \end{itemize}
\textit{(1)} \textit{Police-affiliated} (\(G_{PA}\)): individuals with prior law enforcement experience; \textit{(2)} \textit{Justice-system-impacted} (\(G_{JI}\)): individuals with a history of incarceration or arrest; and \textit{(3)} \textit{Non-affiliated} (\(G_{NA}\)): community members without prior law enforcement experience or history of incarceration/arrest.
This design ensures that the subjectivity motivating our study is reflected in the collected respect annotations, rather than collapsing judgments into a single perspective.

\subsection{Annotation Phases and Guideline Development} \label{sec:phases}
% The annotation framework was developed through iterative refinement informed by %the fieldwork in Section~\ref{sec:annotator_setting}, 
% extensive fieldwork to ensure the schema reflects community expectations and institutional constraints \citep{anonymous2024}. The process proceeds in the following stages:
% % \citep{trager2024everydayrespect_phase0}.
The annotation framework was developed through iterative refinement informed by fieldwork to ensure the schema reflects community expectations and institutional constraints \citep{trager2024everydayrespect_phase0}.
% \citep{anonymous2024}.
To operationalize this process, we built a domain-specific, secure annotation platform  (\autoref{app:dataset_details}). The annotation proceeded in the following stages:
\begin{table}[!htbp]
\centering
\small
\setlength{\tabcolsep}{3pt}
\renewcommand{\arraystretch}{1.15}
\begin{tabularx}{\linewidth}{c l c c c c}
\toprule
 & \textbf{Statistic}
 & \(G_{\mathrm{PA}}\)
 & \(G_{\mathrm{NA}}\)
 & \(G_{\mathrm{JI}}\)
 & \textbf{All} \\
\midrule

% ----------- SHARED STATS -----------
\multirow{2}{*}{\rotatebox[origin=c]{90}{}}
 & \#Annotators   & 5 & 20 & 2 & 27 \\
 & \#Annotations  & 300 & 872 & 190 & 1362 \\
\midrule

% ----------- OFFICER STATS -----------
\multirow{3}{*}{\rotatebox[origin=c]{90}{\textbf{Officer}}}
 & Mean(ratings)             & 3.69 & 3.61 & 4.07 & 3.69 \\
 & Std(ratings)                & 1.00 & 0.78 & 0.63 & 0.83 \\
 & Rationale length (tokens) & 51.6 & 42.7 & 39.1 & 44.2 \\
\midrule

% ----------- DRIVER STATS -----------
\multirow{3}{*}{\rotatebox[origin=c]{90}{\textbf{Driver}}}
 &  Mean(ratings)              & 3.97 & 3.65 & 4.03 & 3.77 \\
 & Std(ratings)                & 0.88 & 0.83 & 0.77 & 0.85 \\
 & Rationale length (tokens) & 43.9 & 42.0 & 39.8 & 42.1 \\

\bottomrule
\end{tabularx}
\caption{Phase~2B respect annotation statistics by annotator group and entity.
}
\label{tab:phase2b_annotators}
\vspace{-0.5cm}
\end{table}

\noindent\textit{Preliminary Phases.} In \textit{Phase 0}, annotators corrected noisy WhisperX \citep{bain23_interspeech} transcripts. %automatic speech recognition to ensure reliable text-based analysis. 
Subsequent stages involved tagging individuals and objects (\textit{Phase 1A}), labeling perceived demographics (\textit{Phase 1B}), and tracking emotions alongside objective stop outcomes (\textit{Phase 2A}).

\noindent\textit{Phase 2B: Multi-Annotator Judgments of Respect.} As the central component of this work, annotators produced fine-grained subjective judgments of respect ratings and rationales for both officers and civilians for a subset of traffic stops. Each sampled video is labeled by one to four annotators with varied police-related lived experiences, enabling the analysis of perceptual %convergence and divergence. 
differences. Find summary statistics of this phase annotations in \autoref{tab:phase2b_annotators}. 

We empirically validate a core assumption underlying this dataset: that annotators’ free-text rationales reflect their corresponding respect ratings for both officers and civilians (see Appendix~\ref{sec:appendix-rating-pred}).

\section{Modeling Subjective Respect Annotations}

We describe our task and the modeling components to predict and explain annotator-specific perceived respect in police–civilian interactions. Our approach relies on a domain-specific rubric for evaluating rationale quality, and uses this rubric to construct preference data for personalized alignment. We begin by defining the task, then present the development and motivation for the rubric, followed by rubric-guided preference data construction.

\subsection{Task Definition}
Given a set of conversations \(C = \{c_i\}_{i=1}^{N}\) from traffic-stop interactions between officers and civilians, and a set of annotators \(A = \{a_j\}_{j=1}^{M}\), we represent the respect annotation provided by annotator \(a_j\) for conversation \(c_i\) as
$y_{ij} = (r_{ij}, \rho_{ij})$,
where \(r_{ij} \in [1,5] \cap \mathbb{Z}\) is the respect rating (from \emph{very disrespectful} to \emph{very respectful}) and \(\rho_{ij}\) is a free-text rationale explaining the annotator’s rating.

Each annotator \(a_j\) is associated with a group label \(g(a_j)\) reflecting their prior exposure to the criminal justice system: \textit{Police-affiliated} (\(G_{PA}\)), \textit{Justice-system-impacted} (\(G_{JI}\)), or \textit{Non-affiliated} (\(G_{NA}\)), as described in \S\ref{sec:annotator_setting}. Each annotator also has a set of demographic attributes \(\mathrm{demo}(a_j)\). These annotator-specific attributes are used in our modeling to signal differences in their perspectives.

We denote the input prompt for conversation \(c_i\), conditioned on annotator \(a_j\), as \(x_{ij}\). The resulting dataset is defined as:
$D = \{d_{ij} : x_{ij} \mapsto y_{ij} \}$
where each data point \(d_{ij}\) maps a prompt to its corresponding rating and rationale.

Our goal is to:
\textit{(1)} predict the overall level of respect \(\hat{r}_{ij}\) displayed by the officer and civilian in conversation \(c_i\), from the perspective of annotator \(a_j\);
\textit{(2)} generate a rationale \(\hat{\rho}_{ij}\) that explains the reason behind the rating provided by annotator \(a_j\), and
\textit{(3)} assess how well the model captures the distinct perspectives of each group \(G_{PA}\), \(G_{JI}\), and \(G_{NA}\) by evaluating performance at the group level.

We formulate this task as a conditional language generation problem. The prompting format used for model training is provided in Appendix~\ref{app:task_prompt}.
% \vspace{-0.1cm}
\subsection{Respect Rubric Development}

Evaluating the subjective nature of perceived respect in traffic stops requires more than predicting a numeric score; it requires understanding the reasoning behind each judgment. For this reason, our annotation process includes a question asking for a free-text rationale accompanying every respect score. These rationales allow us to capture the nuanced cues and diverse perceptions of what constitutes respectful versus disrespectful behavior across different community members.

A central challenge in this task is determining what counts as a \emph{good} rationale. Standard text-similarity metrics (e.g., ROUGE, BLEU) are ill-suited for evaluating respect-related reasoning, as they cannot capture the nuanced, domain-specific cues that annotators rely on. 
More broadly, evaluating rationales in this setting requires a method that compares generated explanations to reference rationales in a structured and interpretable manner. Such an evaluation should be anchored in the specific dimensions of communication relevant to traffic stops, making explicit which elements of respect or disrespect are present, absent, or mischaracterized in a given rationale.
Overall, our evaluation framework needs to satisfy two criteria:
(1) \emph{interpretability and reasoning}: It must reflect the key elements and reasons behind a respect judgment; 
(2) \emph{policing-domain grounding}: It must be explicitly grounded in policing domain expertise and synthesize the full range of respectful and disrespectful communicative behaviors. 

% outlined in our Respect Rubric explained in the following.

\paragraph{Respect Rubric Guideline.}  
To meet these requirements, we develop a survey- and theory-informed guideline that identifies the concrete aspects of officer-civilian communication that annotators commonly reference when explaining their provided respect ratings.
 
The rubric emerges from the following complementary sources: 
\textit{(i)} the annotation manual, which already links objective and subjective variables to prior research; \textit{(ii)} extensive fieldwork%discussed in \S\ref{sec:annotator_setting}
, including the citywide survey, qualitative interviews, researcher ride-alongs, and insights from LAPD training materials; and \textit{(iii)} academic literature on procedural justice, transparency, and de-escalation \citep{sunshine_role_2003, mazerolle_procedural_2013}. Drawing on these sources, the rubric organizes respect into three core (overlapping) categories: \textit{Emotions}, \textit{Professionalism}, and \textit{Communication}. A separate category of \textit{Contextual Moderators} is included to capture situational conditions that can influence the interpretation of the core categories. 

The resulting rubric, shown in Appendix~\ref{app:rubric}, specifies both respectful and disrespectful elements within each category. For example, emotional respect may involve warmth, empathy, or calm body language, while disrespect may manifest as offensiveness or unnecessary anger. Professionalism includes greetings, introductions, and composure under stress, while lapses include abrupt commands or mocking language. Communication covers explanations, comprehension checks, and signaling when civilians are free to leave. Contextual moderators, such as threats or environmental disruptions, shape how these elements are expressed and how annotators justify their ratings. By anchoring rationale evaluation in these descriptors, the rubric allows for systematic, transparent comparison of annotator-written and model-generated reasoning.
% \vspace{-0.4cm}
\paragraph{LLM-as-a-judge.}
We implement a rubric-aware LLM-as-a-judge, validated by human annotators, that maps both ground-truth rationales (\(\rho\)) and model-generated rationales (\(\hat{\rho}\)) to structured representations grounded in the respect rubric. Using \texttt{LLaMA-3-70B} \citep{grattafiori2024llama}, the judge identifies which respect-related elements are present in a given rationale. This rubric-grounded representation enables systematic comparison of human and model reasoning and serves as a shared evaluation primitive throughout our framework. Judge prompts are provided in Appendix~\ref{app:rubric}.

\paragraph{Metric Formulation.}
Let the respect rubric consist of \(K\) dimensions. Given a rationale \(\rho\), the judge produces a binary activation vector
\(
z(\rho) \in \{0,1\}^{K},
\)
indicating the presence or absence of each rubric element.

%To evaluate a generated rationale \(\hat{\rho}\), we compare its activation vector to that of the corresponding reference rationale \(\rho\). We compute precision, recall, and \(F_1\) between the two vectors, denoted \(P(\hat{\rho})\), \(R(\hat{\rho})\), and \(F_1(\hat{\rho})\). 
We evaluate a generated rationale \(\hat{\rho}\) by computing macro precision \(P(\hat{\rho})\), recall \(R(\hat{\rho})\), and \(F_1\) between its activation vector and the reference \(z(\rho)\). 
This rubric-based metric provides an interpretable measure of rationale quality and is used both for evaluation and for constructing preference data in the alignment stage.

% To evaluate a generated rationale \(\hat{\rho}\), we compare its vector representation to that of the corresponding ground-truth rationale \(\rho\). \(Precision\), \(recall\), and \(F_1\) are computed between the two vectors: $F_1\big(z(\hat{\rho}), z(\rho)\big)$.

\subsection{Rubric-Grounded Preference Data Synthesis }
\label{sec:pref_pair_method}

Rationales in this domain are nuanced, structurally varied, and context-dependent.
To align models toward such reasoning, we introduce a rubric-grounded preference dataset synthesis framework composed of the following modules (see \autoref{fig:main}).
% \vspace{-1.8em}
\paragraph{\textsc{Generator Module.}}
Given an input prompt \(x_{ij}\) and a target model \(\pi_t\), the Generator Module produces a set of candidate rationales,
\(
\{\hat{\rho}_{ij}^{(g)}\}_{g=1}^{G}
\sim \pi_t(\cdot \mid x_{ij}).
\)
These candidates may originate from different target models, including a supervised fine-tuned model or a larger base model.
% \vspace{-0.5em}
\paragraph{\textsc{Augmenter Module.}}
To augment high-quality rationales, we apply a paraphrasing model \(\pi_{\mathrm{phr}}\) to the reference rationale \(\rho_{ij}\),
\(
\{\hat{\rho}_{ij}^{(p)}\}_{p=1}^{P}
\sim \pi_{\mathrm{phr}}(\cdot \mid \rho_{ij}),
\)
producing paraphrased variants that preserve the underlying reasoning. These paraphrases are evaluated in the same manner as generated candidates and may be included in the chosen pool if they satisfy the rubric-alignment criteria.

% \vspace{-0.4em}
\paragraph{\textsc{Judge Module.}}
We reuse the rubric-aware judge
% \(\mathcal{J}\)
to evaluate both generated candidates and reference rationales within the preference synthesis pipeline.
For each candidate, \(\hat{\rho}_{ij}^{(g)}\) or \(\hat{\rho}_{ij}^{(p)}\), and the corresponding reference rationale \(\rho_{ij}\), the judge assigns binary rubric activations \(z(\hat{\rho}_{ij})\) and \(z(\rho_{ij})\).

Candidate rationales are assigned to \emph{chosen} or \emph{rejected} pools based on rubric-based alignment with the reference rationale.
For candidates derived from generator module \(\hat{\rho}_{ij}^{(g)}\), we apply empirically determined thresholds, $\tau_{ch}$ and $\tau_{rej}$, to ensure clear separation between high- and low-quality reasoning. A candidate is assigned to the \emph{chosen} pool if
\(
F_1(\hat{\rho}) \geq \tau_{ch}.
\)
A candidate is assigned to the \emph{rejected} pool if it satisfies any rejection condition, including
\textit{(i)} low overall alignment (\(F_1(\hat{\rho}) < \tau_{rej}\)),
\textit{(ii)} low precision or recall (\(P(\hat{\rho}) < \tau_{rej}\) or \(R(\hat{\rho}) < \tau_{rej}\)),
\textit{(iii)} systematic false positives on commonly over-generated rubric dimensions (e.g., \emph{warmth}), or
\textit{(iv)} false negatives on sparse but critical rubric dimensions.
Candidates that satisfy neither condition are discarded.

For $\hat{\rho}^{(p)}$, we apply only the \emph{chosen} criteria: paraphrases that meet the alignment threshold are included in the chosen pool, and others are discarded.

\paragraph{Preference Pair Construction.}
For each instance \(d_{ij}\), we construct up to \(k\) preference pairs by pairing elements from the chosen---which also includes ground truths---and rejected pools:
\(
(x_{ij}, y_{ij}^{+}, y_{ij}^{-}).
\) The resulting preference pairs dataset is used for downstream 
preference optimization of the target model to make the model better aligned with the rubric-grounded respect values embedded in the ground truth respect annotations.

\section{Experiments}
We first present a mixed-effects analysis to quantify how lived experience drives divergent interpretations of respect. We then describe our experiments, which proceed in three stages: \textit{(i)} prompt-level, \textit{(ii)} model-level, and \textit{(iii)} alignment-level.

\subsection{Mixed-Effects Analysis of Respect Annotations.} \label{sec:subj_proof}
A core motivation for our modeling framework is the substantial subjectivity with which annotators perceive respect in police-civilian encounters. As views vary with lived experience, we first examine how annotator background including group identity \(g(a_j)\) and demographic attributes \(\text{demo}(a_j)\), explain variation in the ratings \(r_{ij}\) they assign. To quantify this variation, we fit mixed-effects regression models (details in Appendix~\ref{app:subjectivity-in-respect-annotations}), treating annotators as random effects and incorporating annotator background as predictor variables. 

% Full attribute distributions appear in the Appendix.

Across both officer-respect and civilian-respect models, we find annotator group identity $g(a_j)$ as the only annotator-level attribute that significantly predicted perceived respect. Specifically, annotator group identity has a significant effect on respect annotations of officers (\(\beta = 0.541, p < .001\)) as well as on respect annotations of drivers (\(\beta = 0.501, p = .046\)). In contrast, other annotator demographic variables, including age, gender, and race, do not exhibit significant main effects on respect judgments. Taken together, these results suggest that lived experience with the criminal justice system is the dominant lens through which annotators interpret respect. Additional model specifications and coefficients are included in the Appendix~\ref{app:subjectivity-in-respect-annotations}.

% \subsection{Subjective Respect Prediction Experiments}

\subsection{Prompt-level Experiments}
This stage serves two goals: establishing baseline performance, and assessing whether explicitly conditioning on annotator background information improves subjective predictions. To this end, we augment the task prompt with different subsets of the annotator background variables and evaluate the zero-shot performance of the base model on both rating and rationale generation.

This allows us to test whether making the annotator’s perspective explicit via prompting enables the model to approximate annotator \(a_j\)'s subjective interpretation \emph{without} any parameter updates.

As confirmed by our findings in \hyperref[sec:subj_proof]{Section~\ref*{sec:subj_proof}}, $g(a_j)$ is the strongest predictor of subjective rating variation. Hence, we begin with conditioning the model on \(g(a_j)\) and incrementally incorporate additional background variables. The prompt configurations (examples in Appendix~\ref{app:task_prompt}) are:
\textit{(i)} \(Base\): No annotator information included; 
\textit{(ii)} \(Base^{(g)}\): Group-Personalized Baseline; the prompt is conditioned only on the annotator’s group identity \(g(a_j)\);
 \textit{(iii)} \(Base^{(g+demo)}\): Annotator-Personalized Baseline; the prompt includes group identity and annotator demographic attributes, including age, race, and gender; and
 \textit{(iv)} \(Base^{(g+demo+ent)}\): Target Entity-Personalized Baseline; the prompt includes annotator group, annotator demographics, and perceived demographic attributes of the officer or civilian involved in the conversation.

\subsection{Model-level Experiments}

To adapt the model to the domain and learn personalized generation patterns, we apply parameter-efficient fine-tuning via LoRA~\citep{hu2022lora}. We condition the model on the \((g + demo)\), since this configuration yields better result in zero-shot setting. The prompt template is shown in Appendix~\ref{app:task_prompt}.

During training, we apply the standard autoregressive language modeling objective, computing the negative log-likelihood loss only over the tokens corresponding to the rating \(r_{ij}\) and the rationale \(\rho_{ij}\).
This produces the supervised model \(\pi_{\text{sft}}\) that captures annotator-specific patterns directly from data, enabling it to more accurately predict subjective ratings and produce rationales aligned with annotator \(a_j\)'s interpretive perspective.

\begin{figure*}[!htbp]
    \centering
\includegraphics[width=0.95\linewidth]{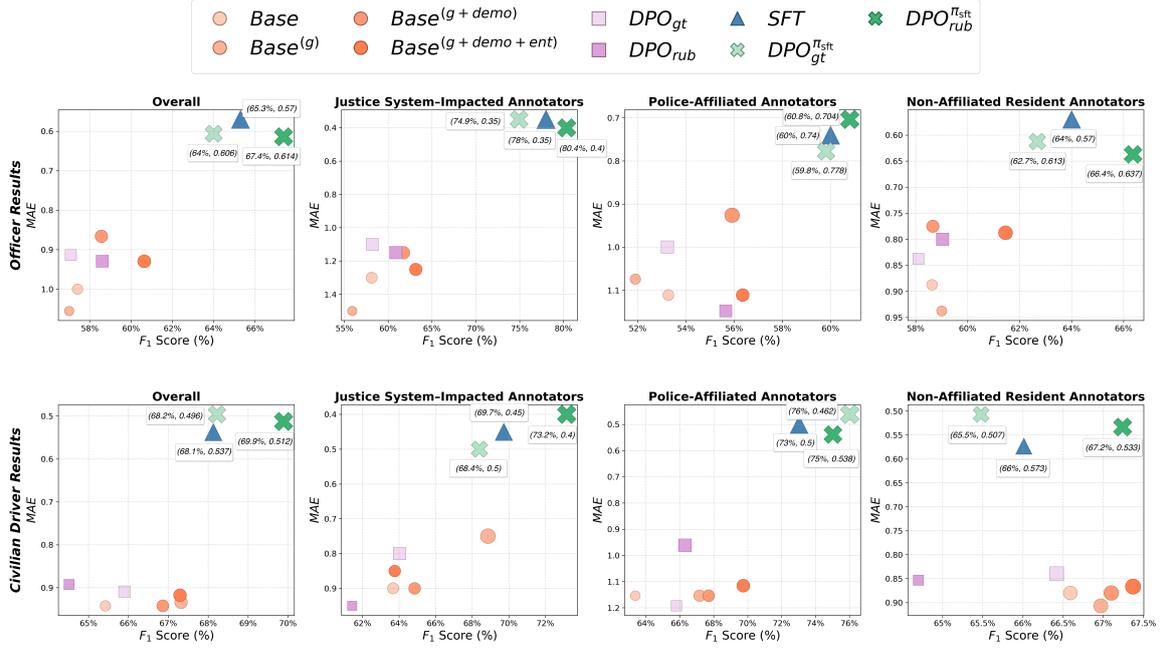}
    \caption{
 Rating MAE (lower is better) 
        and $F_1(\hat{\rho})$ (higher is better) for officers (top) and 
        drivers (bottom).
    }
    \label{fig:combined_mae_f1_scatter}
    % \vspace{-0.3cm}
\end{figure*}

\subsection{Alignment-level Experiments}
To evaluate the impact of alignment, we compare two types of preference datasets: \textit{(i)} pairs derived from the original human annotations (Ground-truth-based)  and \textit{(ii)} augmented rubric-grounded preference data as explained in Section~\ref{sec:pref_pair_method}.
% \subsubsection{Preference Data}
\paragraph{Ground-truth-based Preference Data.}
For this dataset, we construct preference pairs directly from human annotations. For each annotated instance \(d_{ij}\), we define the \emph{chosen output} as the annotator's  rating and rationale, \(y^{+}_{ij} = (r_{ij}, \rho_{ij})\). The \emph{rejected outputs} are the ratings and rationales assigned to the same instance \(d_{ij}\) by other annotators,  denoted \(y^{-}_{i\ell} = (r_{i\ell}, \rho_{i\ell})\) for all \(\ell \neq j\). This process yields preference tuples of the form \((x_{ij}, y^{+}_{ij}, y^{-}_{i\ell})\), capturing natural disagreement across annotators and ensuring that each annotator's perspective is preferred in the alignment objective.

\paragraph{Rubric-Grounded Preference Data.}
 Rubric-driven preference pairs, introduced in Section~\ref{sec:pref_pair_method}, are constructed by contrasting \emph{chosen} high-quality rationales with \emph{rejected} rationales, according to the rubric-based criteria (see Appendix~\ref{app:pref_pairs} for details). Candidate rationales are generated either by the target supervised fine-tuned model \(\pi_{\text{sft}}\) or by a larger language model in a zero-shot setting, specifically \texttt{Qwen3-30B-A3B-Instruct-2507-FP8} \citep{yang2025qwen3}.
 Training the aligned model on this dataset encourages it to generate more specific reasoning, better capture sparse but domain-critical disrespect dimensions, and correct systematic errors in its explanations, thereby moving beyond generic or overly positive statements.

We apply Direct Preference Optimization~\citep[DPO; ][]{rafailov2023direct}
 using both preference datasets and evaluate four alignment configurations:
\textit{(i)} \(DPO_{gt}\): DPO with original data; DPO applied directly to the base model using the original annotation-derived preference pairs.
\textit{(ii)} \(DPO_{rub}\): DPO with rubric-augmented data; DPO applied to the base model using the rubric-constructed preference pairs described above.
\textit{(iii)} \(DPO_{gt}^{\pi_{sft}}\): DPO with original data post-SFT; DPO applied to the supervised model \(\pi_{\text{sft}}\) using the original annotation-derived preference pairs.
\textit{(iv)} \DPOSFTaug{}: DPO with rubric-augmented data post-SFT; DPO applied to \(\pi_{sft}\) using rubric grounded preference data.

This setup enables a controlled and systematic assessment of the impact of each modeling component, allowing us to isolate and attribute the performance gains, particularly those driven by our rubric-based alignment. The implementation details can be found in Appendix~\ref{app:imp_detals}.

\section{Results}\label{sec:results}

We evaluate our models for the two primary interactants in traffic stops: the \emph{officer} and the \emph{driver}. For rating prediction, we report mean absolute error (\(\mathrm{MAE}\); lower is better). For rationale generation, we report rubric-based \fone; higher is better. Results are shown in Figure~\ref{fig:combined_mae_f1_scatter}, aggregated over the test set and per annotator group. 

\subsection{Overall Performance} 

Across both officer and driver evaluations, \DPOSFTaug{} achieves the highest overall \fone.
Zero-shot models perform poorly along both dimensions, exhibiting higher \(\mathrm{MAE}\) and lower \fone, and applying DPO without prior \SFT ~does not yield consistent improvements. In contrast, \SFT{} yields substantial gains, which we further build upon by applying DPO with our constructed datasets (\DPOSFTaug{} and \DPOSFTgt{}) consistently improving \fone over SFT alone, while maintaining rating \(\mathrm{MAE}\) on par with the SFT model. Specifically for officer rationales, the \fone ~improves from $65.3\%$ to $67.4\%$ ($\uparrow2.1\%$), and for driver rationales from $68.1\%$ to $69.9\%$ ($\uparrow1.8\%$).

\subsection{Group-Specific Performance}
We analyze annotator group-specific performance to assess how our alignment method differentially benefit annotators with distinct perspectives. For each group, we report performance on officer and driver respect evaluations, focusing on both rationale quality and rating error.

\noindent\textbf{Justice-System-Impacted Annotators.}
Alignment over \SFT{} improves the performance most strongly for \emph{Justice-system-impacted} annotators. Both \SFT{} and \DPOSFTgt{} achieve the lowest rating error, with \(\mathrm{MAE}\) of 0.35. Additionally, \DPOSFTaug{} improves rationale quality over \SFT{}, increasing \(F_1(\hat{\rho}_{\pi})\) from \(78.0\%\) to \(80.4\%\) (\(+2.4\%\)). For driver, \DPOSFTaug{} achieves the best overall performance, attaining an $\mathrm{MAE}$ of \(0.40\) and improving rationale quality to \(73.2\%\), a gain of \(+3.5\%\) over \SFT{}.
 
\noindent\textbf{Police-Affiliated Annotators.}
For this group, preference-based alignment yields consistent improvements in rationale quality for both officer and driver evaluations. 
For officer respect, \SFT{} achieves an \fone{} of \(60.0\%\) with a corresponding \(\mathrm{MAE}\) of \(0.74\). Applying \DPOSFTaug{} improves \(F_1(\hat{\rho_{\pi}})\) to \(60.8\%\) while reducing rating error to \(0.704\). 
For driver respect, gains from preference alignment are more pronounced. \SFT{} attains an \fone{} of \(73\%\) with \(\mathrm{MAE}_{\text{rating}} = 0.50\), while \DPOSFTgt{} improves rationale quality to \(76\%\) and reduces rating error to \(0.462\).

\noindent\textbf{Non-Affiliated Resident Annotators.}
For officer respect, \SFT{} attains an \fone{} of \(64.0\%\) and \(\mathrm{MAE}\) of \(0.57\). Applying \DPOSFTaug{} further improves the \fone{} to \(66.4\%\) but results in higher rating error (\(\mathrm{MAE} = 0.637\)).
For driver respect, \DPOSFTaug{} performs better relative to \SFT{} alone for both rating and rationale. However, the baseline models perform on par with \DPOSFTaug{} on rationale generation, indicating limited benefit from alignment for this group’s rationale generation. Nevertheless, \DPOSFTaug{}, achieves a lower rating error than the best-performing baseline (\(\downarrow 0.3\)).

% Taken together, these results indicate that while \SFT{} provides a strong foundation, particularly for rating prediction, rubric-grounded preference alignment plays a critical role in improving perspective-specific rationale generation. The largest relative benefits arise for justice system--impacted annotators, followed by police-affiliated annotators, highlighting the value of aligning the rationales to distinct perspectives.

Taken together, these results indicate that while \SFT{} provides a strong foundation—particularly for rating prediction—rubric-grounded preference alignment is crucial for improving perspective-specific rationale generation. The largest relative gains from post-\SFT{} alignment arise for justice-system-impacted annotators, followed by police-affiliated annotators, highlighting the value of aligning rationales to distinct community perspectives.

\section{Related Work}
\noindent\textbf{Respect and Communication in Policing.}  
Research shows that officer respect during traffic stops significantly influences public trust and perceived legitimacy \citep{tyler1988procedural,tyler2017procedural,worden2018measuring, sunshine2003role, nagin2017procedural,jackson2012people}.  These interactions shape broader relations with the state. Black drivers systematically receive less respect in routine stops \citep{voigt2017language}, a reality that likely shapes community expectations and demands for respectful treatment \citep{camp2024leveraging}. Recent research uses BWC transcripts to evaluate officer training on respectful traffic stops  \citep{camp2024leveraging}, as well as  how civilian demeanor affects officer behavior \citep{sunde2023revisiting}. 

\noindent\textbf{Personalization in LLMs.}  
Personalization is a major focus in LLM research, with surveys detailing its evolution. Input-level methods \citep{liu2025survey} encode identity via persona-guided prompts \citep{ryan-etal-2025-synthesizeme}, embeddings \citep{li-liang-2021-prefix,doddapaneni-etal-2024-user,huber2025embedding,liu-etal-2025-llms}, soft adapters \citep{hebert2024persoma}, or distilled prompts \citep{ramos2024peapod}, while conversational systems generate tailored narratives \citep{sayana2025beyond}. Model-level approaches utilize parameter-efficient adaptation, such as personalized LoRA variants for various tasks \citep{zhang2024personalized,tan-etal-2024-personalized,zhu2024lifelong, kong2024customizing,long2024dual} and personality customization via mixture of experts \citep{dan2025p}.
Alignment-based strategies modify objectives through group optimization \citep{zhao2023group}, personalized reinforcement learning \citep{li2024personalized}, or parameter merging \citep{jang2023personalized}, alongside inference-time steering for retraining-free control \citep{he2024context,cao2024personalized,bo2025steerable,zhang-etal-2025-personalized}. Recent advances include memory-based systems \citep{zhang2025prime}, latent difference modeling \citep{qiu-etal-2025-latent, qiu_etal_2025_measuring}, causal alignment \citep{zhao2025nextquill}, and lifelong adaptation \citep{wang2024ai}.  

\noindent\textbf{Reinforcement Learning with AI Feedback.}  
Aligning models for personalization relies on human feedback \citep{ouyang2022training}, yet scaling diverse alignment data is challenging. Reinforcement Learning with AI Feedback (RLAIF) addresses this by using LLMs to generate and rank preference pairs \citep{bai2022constitutional,lee2024rlaif}, though it often suffers from low response diversity. Strategies to enhance diversity include using ensembles of different model families \citep{cui2024ultrafeedback}, iterative prompt and response refinement \citep{dongself}, and policy gradient-based prompt updates \citep{zhou2025anyprefer}. While RLAIF typically ranks candidates via specific criteria \citep{yuan2024self} or generation probabilities \citep{lee2024rlaif}, standard Likert-style scoring often yields inconsistent judgments \citep{lee-etal-2025-checkeval}. To improve reliability, recent work proposes rubric-based evaluations where models answer targeted binary questions across multiple dimensions \citep{lee-etal-2025-checkeval,wei2025rocketeval,cook2024ticking,ruan2025expertlongbench}.

\section{Conclusion}
%\vspace{-5pt}
This work introduces large-scale LAPD BWC annotations that capture the heterogeneous ways diverse community members interpret officer and civilian behavior. By modeling respect as an inherently subjective task, we move beyond a singular ground-truth to incorporate the divergent perspectives of justice-system-impacted, police-affiliated, and other community annotators. We develop a domain-specific rubric grounded in community insights and a framework combining SFT with DPO. By constructing preference pairs via rubric alignment, our approach mitigates reasoning failures and improves rating accuracy and rationale quality.

Our results show that incorporating group-specific context is essential for effective alignment. Gains are most significant for justice-system-impacted annotators, highlighting the necessity of modeling divergent perspectives. Overall, this work demonstrates a framework for aligning language models to nuanced, explanation-dependent judgments, offering a pathway toward perspective-aware AI systems in high-stakes social domains.

\section{Limitations}
We acknowledge that while our dataset includes annotators from multiple stakeholder groups with diverse lived experiences, annotator diversity can always be expanded. In particular, additional demographic, geographic, and experiential perspectives, both within and beyond the groups considered here, may further enrich the range of interpretations captured and improve the robustness of personalized modeling.
Second, our analysis focuses exclusively on the textual modality derived from body-worn camera transcripts. Although language is a central channel through which respect is communicated, nonverbal cues such as tone, prosody, facial expressions, and physical positioning also play a critical role in shaping perceptions of respect in police--civilian interactions. Incorporating audio and visual modalities remains an important avenue for our future work.
Finally, our modeling and evaluation framework is grounded in a domain-specific respect rubric developed for traffic-stop encounters. While this rubric enables structured and interpretable reasoning in this context, its dimensions and criteria may not directly transfer to other forms of police--community interactions or to different institutional settings without adaptation. Extending this approach to additional domains will require careful reconsideration of rubric design and stakeholder perspectives.

\section{Ethical Statement}
\paragraph{Bias Amplification and Reification of Group Categories.} Conditioning on annotator group identity enables modeling of subjective variation but risks reifying coarse group labels or reinforcing stereotypes if treated as essential. We emphasize that these groups serve as analytic proxies for lived experience, not fixed representations, and that substantial within-group heterogeneity remains.

\paragraph{Fairness and Differential Stakeholder Impact.}
By engaging diverse stakeholders and reflecting their perspectives in the AI tools we develop, we are working explicitly to increase fairness and stakeholder representation in this field.

\paragraph{Privacy and Data Sensitivity.}
 This work relies on sensitive police body-worn camera data. All annotations were conducted within secure, on-premises infrastructure, and we obtain informed consent from annotators to study their annotations. We paid annotators wages ranging from \$17.28 per hour for undergraduate students to \$30 per hour for more experienced professionals.  Wages were based on experience level. Consistent with hiring practices at [university anonymized], annotators were recruited through a range of measures, including the posting of job listings on a range of public platforms.
The study was approved by the IRB Board at [university anonymized].

 Data sharing is strictly constrained by contractual agreements with LAPD. We do not release raw video, audio, or unredacted transcripts, and any disseminated artifacts are anonymized and access-controlled. Despite these safeguards, modeling real-world interactions involving vulnerable individuals carries inherent privacy risks.

\paragraph{Safeguards and Mitigation Strategies.}
We take several steps to mitigate these risks: (1) framing respect as inherently subjective and rejecting a single ground truth; (2) evaluating models using rubric-based, interpretable criteria rather than opaque scores; (3) emphasizing human-in-the-loop analysis rather than automated decision-making; and (4) limiting claims about deployment. Future work could explore gated release mechanisms and participatory governance with community stakeholders.
\bibliography{custom}

\appendix
\section{Dataset details} \label{app:dataset_details}
The sampled stops statistics is shown in Table~\ref{tab:stops_stats}.
To operationalize the annotation process, we design a domain-specific, secure, on-premises annotation software and pipeline. This custom solution is necessitated both by the complex task demands and by strict security constraints that prohibited the raw BWV data from leaving LAPD headquarters, where all annotations are completed \citep{hejabi2024cvat}.
% \citep{anonymous2024}. Figure~\ref{fig:anot_plat} shows our annotation platform.

\subsection{Preference Pairs} \label{app:pref_pairs}

We construct preference pairs using \textit{generator} and \textit{augmenter} modules as introduced in the main paper, which leverages ground-truth rationales, and their paraphrased variants as chosen samples. For paraphrasing, we employ \texttt{Qwen/Qwen3-Next-80B-A3B-Instruct} \citep{yang2025qwen3} with temperature $0.7$, top-$p$ $=1.0$, and a maximum generation length of $2000$ tokens to generate three semantically equivalent paraphrases for each ground-truth rationale while preserving the original respect rating and rubric-dimensional structure. The paraphrasing prompt is shown in Figure~\ref{fig:paraphrase_prompt}. Chosen samples consist of the original ground-truth rationales and their paraphrased versions, ensuring that all chosen samples maintain perfect rating accuracy.
Rejected samples are selected based on the rubric-alignment quality metrics described in the Section~\ref{sec:pref_pair_method}. Specifically, we reject model-generated rationales that exhibit any of the following failure modes: (i) binary precision below $0.5$, (ii) binary recall below $0.5$, or (iii) false-positive activations on target emotional-respect dimensions, namely \texttt{emotional\_respect.respectful.empathy} or \texttt{emotional\_respect.respectful.warmth}. For each chosen sample (ground truth or paraphrase), we construct up to $5$ preference pairs by randomly sampling from the pool of valid rejected candidates, enforcing a minimum rubric $F_1$ score gap of $0.2$ between chosen and rejected outputs. These thresholds and pairing constraints correspond to the final configuration selected after empirically evaluating multiple alternative settings, and together result in a dataset of $10{,}612$ rubric-grounded preference pairs.
This approach results in a dataset where chosen samples demonstrate high alignment with ground truth annotations (near-perfect $F_1$ for paraphrases), while rejected samples demonstrate substantially lower alignment, with average precision of $0.41$ and recall of $0.51$, creating clear preference signals for alignment training.

\setlength{\tabcolsep}{2pt}
\begin{table}[h]
\centering
\small
\begin{tabular}{lr}
\toprule
\textbf{Stops Statistic} & \textbf{Count (\%)} \\
\midrule 
Number of stops                & 1008 (--) \\ 
Timeframe of stops             & -- \\ 
Total duration (hours)         & 696.7 (--) \\ 
Mean duration (minutes)        & 14.9 (SD = 13.7) \\ 
\midrule
\multicolumn{2}{l}{\textit{Reason}} \\
\quad Traffic Violation                    & 990 (82.8\%) \\
\quad Reasonable suspicion of crime        & 173 (14.5\%) \\
\quad Parole/Probation                     & 16 (1.3\%) \\
\quad Arrest Warrant/Wanted Person         & 14 (1.2\%) \\
\quad Consensual encounter w/search        & 2 (0.2\%) \\
\quad Possible Danger to Self\&Others/5150 & 1 (0.1\%) \\
\midrule
\multicolumn{2}{l}{\textit{Action taken}} \\
\quad Search of person conducted           & 520 (43.5\%) \\
\quad None                                 & 452 (37.8\%) \\
\quad Handcuffed/flex cuffed               & 415 (34.7\%) \\
\quad Search of property conducted         & 393 (32.9\%) \\
\quad Ordered from vehicle                 & 361 (30.2\%) \\
\quad Curbside detention                   & 265 (22.2\%) \\
\quad Req. consent to search property      & 146 (12.2\%) \\
\quad Patrol car detention                 & 131 (11.0\%) \\
\quad Req. consent to search person        & 79 (6.6\%) \\
\quad Vehicle impounded                    & 67 (5.6\%) \\
\quad Firearm pointed at person            & 40 (3.3\%) \\
\quad Field sobriety test                  & 32 (2.7\%) \\
\quad Property was seized                  & 31 (2.6\%) \\
\quad Person photographed                  & 9 (0.8\%) \\
\quad Physically removed from vehicle      & 3 (0.3\%) \\
\midrule
\multicolumn{2}{l}{\textit{Result}} \\
\quad Warning                              & 369 (30.9\%) \\
\quad Citation for infraction              & 348 (29.1\%) \\
\quad FI card completed                    & 236 (19.7\%) \\
\quad No action                            & 207 (17.3\%) \\
\quad Arrest w/o warrant                  & 164 (13.7\%) \\
\quad Warrant arrest                      & 24 (2.0\%) \\
\quad In-field cite and release            & 8 (0.7\%) \\
\quad Psychiatric hold                    & 3 (0.3\%) \\
\bottomrule
\end{tabular}
\caption{Summary statistics of the LAPD traffic stop videos.}
\label{tab:stops_stats}
\end{table}

\section{Respect Rubric} \label{app:rubric}
The prompts used for the LLM-as-a-judge to evaluate the rationales based on the respect rubric for officer and driver are shown in Figures~\ref{fig:officer_rubric} and \ref{fig:drvier_rubric} respectively.

\section{Task prompts} \label{app:task_prompt}
Example of the prompt used for training and evaluation of our models is shown in Figure~\ref{fig:task_prompt}

\section{Subjectivity in Respect Annotations}
\label{app:subjectivity-in-respect-annotations}
To investigate the subjectivity inherent in respect ratings during police stops, we employed mixed-effect models that incorporate both fixed and random effects. These models assess how annotator-related variables contribute to variation in respect ratings. Our primary focus is on the identity and experience of the annotators, particularly whether they have previously been incarcerated or affiliated with law enforcement, as well as their demographics (race, gender, age). We also include perceived demographic and socioeconomic attributes of the drivers and officers, such as race, gender, age, height, presence of a foreign accent, clothing, and car type, as control variables to account for potential confounding influences and to ensure that the estimated effects of annotator group identity are not driven by differences in perceived characteristics of the individuals involved. 

% \begin{figure}[t]
%     \centering
%     \includegraphics[width=\columnwidth]{figures/respect_distribution_by_annotator_group.png}
%     \caption{Distribution of respect ratings across annotator groups.}
%     \label{fig:respect-ratings-differences-annotator-group}
% \end{figure}

\paragraph{Police officers.}
Annotator group identity was the single significant predictor of perceived officer respect among annotator background attributes. Justice system-impacted annotators rated officers as more respectful than non-affiliated annotators (\(\beta = 0.553, p = .002, CI = [0.206, 0.900]\)), and police-affiliated annotators also provided higher officer-respect ratings than non-affiliated annotators (\(\beta = 0.526, p = .007, CI = [0.141, 0.911]\)). No other annotator-level demographic variables (e.g., race, gender, age) significantly influenced judgments, underscoring that lived experience with the criminal justice system was the primary factor shaping evaluations of police behavior.

Controlling for perceived officer and driver characteristics did not change the observed effect of annotator group on respect annotations. Overall, officers were judged as less respectful when they were perceived to have a foreign accent (\(\beta = -1.358, p < .001, CI = [-2.031, -0.685]\)). Several driver attributes also shaped how respectful the officer’s behavior appeared: officers were rated as more respectful when the driver was female (\(\beta = 0.282, p = .002, CI = [0.107, 0.458]\)), when the driver was short (\(\beta = 0.397, p = .007, CI = [0.110, 0.684]\)), and when the driver was perceived to be driving a middle-class car, compared to a working-class car (\(\beta = 0.265, p = .020, CI = [0.042, 0.487]\)). Officers were judged as less respectful when the driver spoke with a foreign accent (\(\beta = -0.344, p = .004, CI = [-0.581, -0.108]\)).

% These findings suggest that, while various visible attributes are available to annotators, only a limited set meaningfully shaped their judgments. Nevertheless, annotators' own lived experience with the justice system remained the largest factor in how they interpreted officer respect.

% Altogether, model comparisons consistently favored specifications that included the annotator group and the officer's perceived accent. These results underscore the central role of annotator background in shaping perceived respect, with annotator group membership emerging as the most robust and consistent explanatory variable.

\paragraph{Drivers.}
Similarly, annotator group identity emerged as the single significant predictor of perceived driver respect among annotator background attributes. Police-affiliated annotators rated drivers as significantly more respectful than non-affiliated annotators (\(\beta = 0.837, p = .010, CI = [0.203, 1.470]\)), whereas we found no evidence of a difference between formerly incarcerated annotators and the non-affiliated group (\(\beta = 0.174, p = .588, CI = [-0.457, 0.805]\)). Individual annotator demographics (race, gender, age) were not predictive of the annotators' respect judgment, indicating that group identity remained the primary factor shaping perceptions of driver behavior.

Controlling for perceived officer and driver characteristics did not change the observed effect of annotator group on respect annotations. Overall, a small number of stop-level characteristics also influenced perceived driver respect. Drivers perceived as Hispanic were rated as more respectful than those perceived as White (\(\beta = 0.320, p = .012, CI = [0.071, 0.569]\)). Several officer-related attributes also shaped judgments: when the officer was perceived as Asian, drivers were rated as less respectful (\(\beta = -0.728, p = .018, CI = [-1.331, -0.124]\)), and when the officer was perceived to have a foreign accent, drivers were judged as substantially less respectful (\(\beta = -1.513, p < .001, CI = [-2.137, -0.889]\)). 

% Taken together, these results show that although a few situational and demographic variables affect perceived driver respect, their influence is limited in comparison to the annotator's group identity, which remains the most consequential determinant of how respect is interpreted in police-civilian encounters.

% The best-performing models for driver respect again included annotator group identity and perceived driver race. This mirrors the results for officers and reinforces the idea that annotator experience with the justice system is the most influential factor in shaping respect judgments—above and beyond the observable characteristics of the people involved in the stop.

\paragraph{Discussion.}
Across both officer and driver evaluations, annotator group membership, particularly having been incarcerated or affiliated with law enforcement, stood out as the sole annotator-level factor associated with systematic differences in respect annotations. Interestingly, we observed an inverse pattern of expectations: former officers rated drivers as more respectful, while formerly incarcerated individuals rated officers as more respectful, relative to the non-affiliated group. This counters the intuition that shared background or adversarial experience should yield negative evaluations and instead underscores the complex, subjective nature of respect judgments.

In contrast, demographic characteristics of annotators (e.g., race, gender, age) did not significantly predict respect ratings. Incorporating perceived demographic and socioeconomic attributes of officers and drivers as control variables revealed additional context-specific patterns, but did not alter the observed effect of annotator group identity on respect annotations. Together, these findings indicate that experiential background related to the criminal justice system plays a central role in shaping perceptions of respect in police–civilian encounters, beyond what can be explained by demographic characteristics alone.

\section{Implementation Details} \label{app:imp_detals}

We use Qwen2.5-7B-Instruct \citep{qwen2025qwen25technicalreport} as the base model in our experiments. We fine-tune Qwen2.5-7B-Instruct using LoRA, and then, using Direct Preference Optimization (DPO) with parameter-efficient LoRA adapters. Our implementation builds on the HuggingFace \texttt{transformers} and \texttt{trl} libraries, with additional infrastructure for custom checkpoint selection, group-aware inference, and training-time evaluation.
% LoRA adapters (\(r=32\)) are applied to all linear layers, and training is performed in \texttt{bfloat16} precision when supported, otherwise falling back to \texttt{float16}. The DPO reference model is the frozen base model.

Data are split at the conversation level into 80/10/10 train/validation/test partitions to avoid transcript leakage. Fine-tuning is performed with the TRL \texttt{DPOTrainer} using the standard sigmoid DPO loss, optimized with AdamW and a learning rate of \(1\times10^{-7}\). The effective batch size is 32 (batch size 4 with gradient accumulation of~8), and training proceeds for up to 5 epochs with evaluation every 10 steps. All evaluation checkpoints are saved for post-hoc scoring, and the best model is selected based on validation-set rating MAE.

During inference, we use sampling config parameters \texttt{temperature} = 0.5, \texttt{top\_p} = 0.85, and \texttt{top\_k} = 30. All evaluation metrics are computed both overall and per annotator group.

All experiments are conducted on NVIDIA H200 GPUs. We fix the random seed to 42 for reproducibility, and we store hyperparameters, configuration files, and training logs with each checkpoint. Training traces and experiment metadata are logged to Weights \& Biases.

\begin{table}[h!]
\centering
\small
\begin{tabular}{p{1.4cm} p{2.5cm} p{2.2cm}}
\toprule
\textbf{Category} & \textbf{Hyperparameter} & \textbf{Value} \\
\midrule
Base Model & Model & Qwen2.5--7B \\
           & Precision & bf16 / f16 \\
\midrule
Training   & Learning rate & \(1\times10^{-7}\) \\
           & Optimizer & AdamW \\
           & Batch size & 4 \\
           & Grad. accum. & 8 \\
           & Eff. batch size & 32 \\
           & Epochs & 5 \\
           & Eval freq. & Every 10 steps \\
\midrule
LoRA       & Rank & 32 \\
           & Target modules & Linear layers \\
           & \(\alpha\) & 64 \\
\midrule
DPO        & Loss type & Sigmoid \\
           & Beta (\(\beta\)) & 0.1 \\
           & Ref. model & Frozen base \\
           & Truncation & keep\_end \\
\midrule
Data       & Split & 80/10/10 \\
           & Split level & Conversation \\
           & Augment. & Synthetic pairs \\
\bottomrule
\end{tabular}
\caption{Hyperparameters for all experiments.}
\label{tab:hyperparams}
\end{table}

\section{Rationales' Predictiveness of Ratings} \label{sec:appendix-rating-pred}

We empirically validate that the rationales provided by the annotators are actually reflective of their ratings they provide for respect. To do so, we condition LLMs on the rationale, $\rho_{ij}$, and predict the corresponding rating, $r_{ij}$.

Because some of the rationales explicitly mention the ratings, e.g., ``I rate this behavior as \textit{respectful}'', or ``I rated the officer's respect a 4.'', we preprocess the rationales to remove any mention of the label names or their Likert value, and replace these mentions with \texttt{[MUTE]}. In this manner, we focus on the reasons provided for the rating, and therefore use the LLMs as models for the reasoning of the annotators, rather than merely tool to retrieve the rating from the rationale.

We experiment with 5 models, GPT-OSS-20B\footnote{\url{https://huggingface.co/openai/gpt-oss-20b}}, Llama-3.1-8B 8B-Instruct\footnote{\url{https://huggingface.co/meta-llama/llama-3.1-8b-instruct}}, Llama-3.3-70B-Instruct\footnote{\url{https://huggingface.co/meta-llama/llama-3.3-70b-instruct}}, and Qwen3-4B-Instruct\footnote{\url{https://huggingface.co/qwen/qwen3-4b-instruct-2507}} and Qwen3-30B Instruct\footnote{\url{https://huggingface.co/qwen/qwen3-30b-a3b-instruct-2507}}. We set the reasoning effort of the GPT-OSS models to \textit{low}, with a computing budget of 100 tokens. All models are prompted with 5-shot prompts, and each experiment is repeated 5 times, randomly resampling the demonstrations in the prompts from a small held-out training set. Example 1-shot prompts can be seen in Figure~\ref{fig:rr_pred_prompts}.

Table~\ref{tab:ratings-from-rationales} shows the Mean Absolute Error (MAE) with 95\% confidence intervals after converting the predictions of the LLM back to the original 1-5 Likert scale. Note that models are about $60\%$ accurate in predicting the level of respect assigned by the annotator. By looking at their MAE, we see that when they err, they usually predict neighboring values. For reference, we note that without redaction of explicit mentions of the ratings from the rationales, Llama-3.1-8B-Instruct achieved $66.1\%$ accuracy and an MAE of $0.366$, significantly better than with the redacted version. We conclude that rationales capture the valence of the respect effectively.

\begin{table}[!h]
    \centering
    \begin{adjustbox}{width=0.45\textwidth}
    \begin{tabular}{lcc}
        \toprule
        & \multicolumn{2}{c}{MAE} \\
        \cmidrule{2-3}
        \textbf{Model} & \textbf{Officer} & \textbf{Civilian} \\
        \midrule
        GPT-OSS-20B & \rescell{0.547}{0.016} & \rescell{0.529}{0.011}  \\
        % GPT-OSS 120B \\
        Llama-3.1-8B-Instruct & \rescell{0.455}{0.011} & \rescell{0.462}{0.007} \\
        Llama-3.3-70B-Instruct & \rescell{0.404}{0.012} & \rescell{0.424}{0.005} \\
        Qwen3-4B-Instrcut & \rescell{0.459}{0.009} & \rescell{0.449}{0.008} \\
        Qwen3-30B-Instruct & \rescell{0.420}{0.006} & \rescell{0.439}{0.006} \\
        \bottomrule
    \end{tabular}
    \end{adjustbox}
    \caption{Mean Absolute Error (\textit{MAE}) when predicting the respect rating of each annotator based on their provided rationale.}
    \label{tab:ratings-from-rationales}
\end{table}

\begin{figure*}
    \centering
    \includegraphics[width=\textwidth]{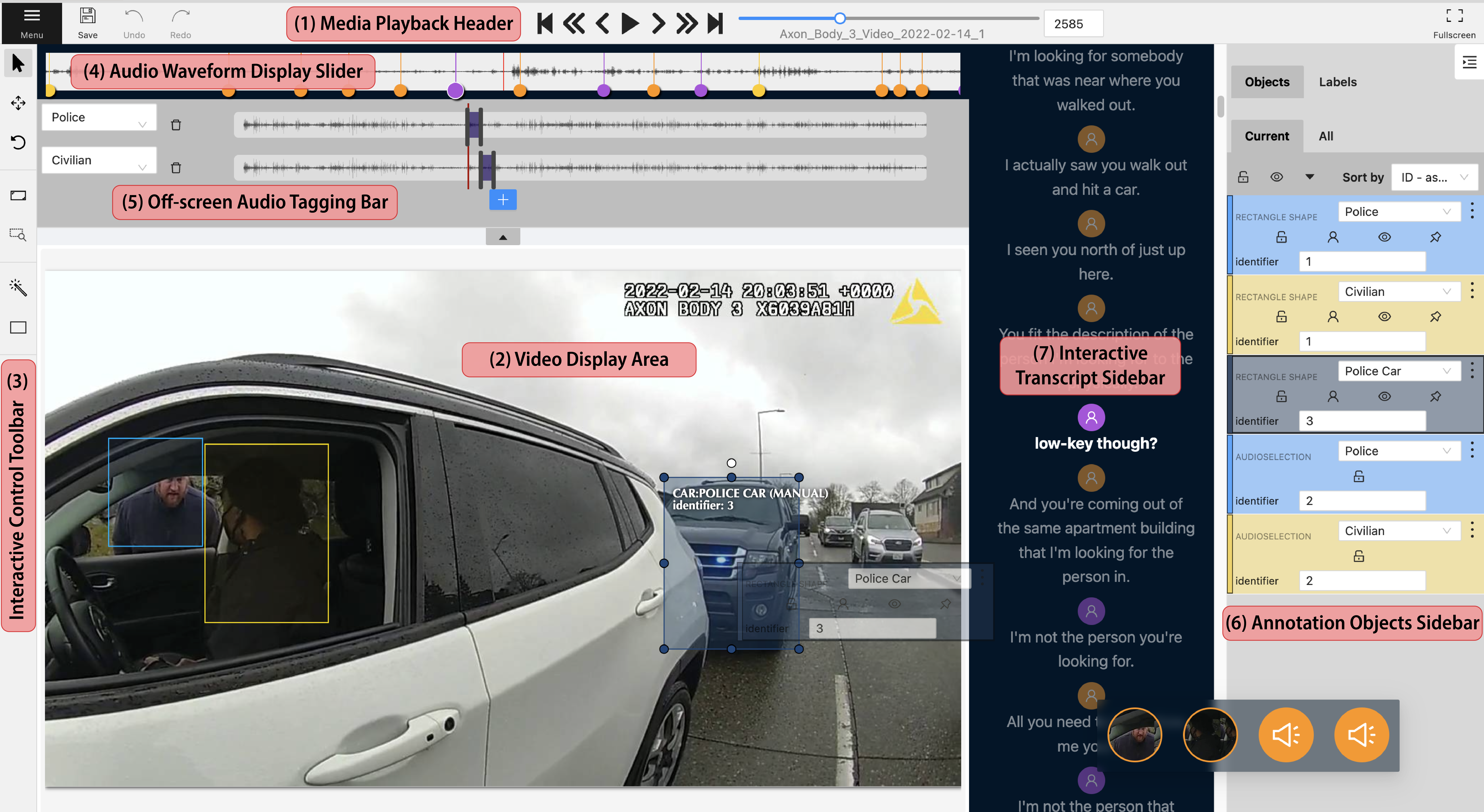}
    \caption{Our annotation platform developed for annotating BWV data.}
    \label{fig:anot_plat}
\end{figure*}

\clearpage
\begin{figure*}[p]
\centering

\begin{subfigure}[t]{0.5\textwidth}
\centering
\begin{tcolorbox}[
 title={Officer respect rubric prompt },
  boxrule=0.6pt,
  arc=1mm,
  left=1mm,right=1mm,top=1.5mm,bottom=1.0mm,
  width=\linewidth,
  % height=0.86\textheight,  % adjust to fit caption
  valign=top
]
\tiny

You are tasked with evaluating rationales explaining ratings of officer respect in police traffic stops. Use the rubric below to guide your evaluation. The rubric describes four major elements of respect: Emotional Respect, Professional Respect, Communicative Respect, and Contextual Moderators of Respect. These elements are not mutually exclusive or hierarchical. Instead, they capture overlapping aspects of officer behavior that annotators may draw on when explaining their ratings.

Your role is to judge whether each of the respect dimensions are referenced in the provded rationale.

\textbf{1. Emotional Respect}

Emotional respect refers to how the officer’s language, tone, and demeanor convey care, warmth, or hostility. It highlights the affective dimension of respectful or disrespectful communication.

\textbf{Respectful Emotional Descriptives}

Warmth :The officer conveys friendliness or kindness.\\
Example: “The officer smiled and used a polite tone.”

Empathy: The officer acknowledges or validates the civilian’s perspective or feelings.\\
Example: “The officer said they understood why the driver was nervous.”

Apology/Thanks: The officer apologizes for inconvenience or expresses gratitude.\\
Example: “They apologized for the delay.”

\textbf{Disrespectful Emotional Descriptives}

Lack of Warmth: The officer is cold, distant, or unfriendly.

Lack of Empathy: The officer doesn't show any understanding or concern.\\
Example: “The officer was not empathetic”

Lack of Apology/Thanks: The officer doesn't apologize or thank.

Offensiveness/Bias: The officer uses insulting, hateful, or biased language.\\
Example: “They implied bias by questioning whether the driver ‘belonged here.’”

Unnecessary Escalation: The officer amplifies tension rather than calming the situation.\\
Example: “They raised their voice aggressively even though the driver was compliant.”

Disrespect for Time: The officer prolongs the stop without clear justification.\\
Example: “The driver was left waiting on the curb for an extended period without explanation.”

% \medskip\hrule\medskip

\textbf{2. Professional Respect}

Professional respect reflects formality, neutrality, and composure. It concerns whether the officer behaves in a manner consistent with professional standards.

\textbf{Respectful Professionalism}

Greeting: The officer begins with a polite greeting rather than a command.

Introduction: The officer introduces themselves or their department.

Professional Language: The officer avoids slang, sarcasm, or mocking tones.\\
Example: “They used clear, formal language.”

Composure/Deflection: The officer maintains calm and redirects tension.\\
Example: “They calmly de-escalated when the driver grew agitated.”

\textbf{Disrespectful Unprofessionalism}

Order Opening: The officer starts with an order instead of a greeting.\\
Example: “They asked for license and registration, without greeting.”

Non-Introduction: The officer fails to identify themselves or their department.\\
Example: “They never told the driver who they were.”

Unprofessional Tone or Language: The officer uses sarcasm, slang, or mockery.\\
Example: “They laughed at the driver’s explanation.”

% \medskip\hrule\medskip

\textbf{3. Communicative Respect}

Communicative respect centers on whether the officer provides transparency, voice, and fairness in the interaction.

\textbf{Respectful Communicative Dialogue}

Reason Ask: The officer invites the civilian to share their perspective.\\
Example: “The officer asked the driver to explain what happened.”

Reason Given: The officer allows the civilian to provide their own account.\\
Example: “The driver explained they were late to pick up their child.”

Explanation: The officer clarifies the reason for the stop or decision.\\
Example: “They explained the traffic law that was violated.”

Options/Next Steps: The officer informs the civilian of their choices.\\
Example: “The officer said you can contest this ticket in court.”

Comprehension Check: The officer ensures the civilian understands.\\
Example: “The officer made sure the driver unserstood why they were pulled over”

Free to Leave: The officer explicitly signals when the encounter is over.\\
Example: “The officer signaled the driver is free to go, and said drive safely.”

\textbf{Disrespectful Communicative Dialogue}

Lack of Reason Ask: The officer doesn’t ask for the civilian’s perspective.

Lack of Reason Given: The officer doesn’t allow explanation.

Lack of Explanation: The officer doesn’t explain the reason for the stop.

Lack of Options/Next Steps: The officer doesn’t inform about choices.

Lack of Comprehension Check: The officer doesn’t verify understanding.

Lack of Free to Leave: The officer doesn’t state that the encounter is over.

Interrupts: The officer cuts off or talks over the civilian.\\
Example: “The driver tried to explain, but the officer repeatedly interrupted.”

% \medskip\hrule\medskip

\textbf{4. Contextual Moderators of Respect}

Contextual moderators are situational factors that shape how respect is expressed or constrained. They are not inherently respectful or disrespectful but help explain why annotators may justify ratings differently.

\textbf{Threat}

Threaten Violence: The civilian makes verbal or physical threats.\\
Example: “The driver threatened to harm the officer.”

Non-Visible Hands: The civilian hides or refuses to show hands.\\
Example: “The officer repeatedly asked to see the driver’s hands.”

Movement Resistance: The civilian resists physical or verbal compliance.\\
Example: “They refused to step out of the vehicle.”

\textbf{Disruptiveness}

Yelling: The civilian shouts over the officer.\\
Example: “The driver kept screaming, preventing any dialogue.”

Extreme Interruptions: The civilian persistently interrupts.\\
Example: “They wouldn’t let the officer finish a sentence.”

Environmental Distractions: Noise or activity in the setting impedes communication.\\
Example: “Passing cars and shouting bystanders made it impossible to hear.”

\end{tcolorbox}
\caption{Part I.}
\end{subfigure}
\hfill
\begin{subfigure}[t]{0.49\textwidth}
\centering
\begin{tcolorbox}[
  boxrule=0.6pt,
  arc=1mm,
left=1.2mm,right=1.2mm,top=1.5mm,bottom=1.0mm,
  width=\linewidth,
  % height=0.86\textheight,
  valign=top
]
\tiny
\setlength{\parindent}{0pt}
\setlength{\parskip}{2pt}
\setlist[itemize]
{leftmargin=1.2em,topsep=1pt,itemsep=1pt,parsep=0pt}

% \medskip\hrule\medskip  

\textbf{RATIONALE TO EVALUATE:}\\
\{rationale\}

% \medskip\hrule\medskip

\textbf{INSTRUCTIONS FOR OUTPUT}

After reading the rationale, evaluate it against the rubric above.\\
For each sub-dimension, answer the following question:\\
Is this aspect being referenced in the rationale, either explicitly or implicitly implied?\\
Respond with ``yes'' or ``no'' for each field.

When answering:
\begin{itemize}
  \item Treat respectful and disrespectful dimensions as separate and independent.
  \item Mark ``yes'' under a respectful field only if the rationale explicitly mentions the presence of that positive behavior.
  \item Mark ``yes'' under a disrespectful field only if the rationale explicitly mentions the absence or violation of that behavior.
  \item Do not mark both as ``yes'' unless the rationale clearly describes both respectful and disrespectful moments.
  \item If the rationale does not reference that aspect at all, mark ``no'' in both.
\end{itemize}

You must produce only valid JSON that conforms exactly to the schema below.

{\tiny
\begin{verbatim}
Schema :
{
  "emotional_respect": {
    "respectful": {
      "warmth": "",
      "empathy": "",
      "apology_thanks": ""
    },
    "disrespectful": {
      "lack_of_warmth": "",
      "lack_of_empathy": "",
      "lack_of_apology_thanks": "",
      "offensiveness_bias": "",
      "unnecessary_escalation": "",
      "disrespect_for_time": ""
    }
  },
  "professional_respect": {
    "respectful": {
      "greeting": "",
      "introduction": "",
      "professional_language": "",
      "composure_deflection": ""
    },
    "disrespectful": {
      "order_opening": "",
      "non_introduction": "",
      "unprofessional_tone_language": ""
    }
  },
  "communicative_respect": {
    "respectful": {
      "reason_ask": "",
      "reason_given": "",
      "explanation": "",
      "options_next_steps": "",
      "comprehension_check": "",
      "free_to_leave": ""
    },
    "disrespectful": {
      "lack_of_reason_ask": "",
      "lack_of_reason_given": "",
      "lack_of_explanation": "",
      "lack_of_options_next_steps": "",
      "lack_of_comprehension_check": "",
      "lack_of_free_to_leave": "",
      "interrupts": ""
    }
  },
  "contextual_moderators": {
    "threat": {
      "threaten_violence": "",
      "non_visible_hands": "",
      "movement_resistance": ""
    },
    "disruptiveness": {
      "yelling": "",
      "extreme_interruptions": "",
      "environmental_distractions": ""
    }
  }
}
\end{verbatim}
}

\end{tcolorbox}
\caption{Part II.}
\end{subfigure}
\caption{Rubric and instructions used by the LLM-as-a-judge to evaluate officer-respect rationales and output structured rubric activations.}
\label{fig:officer_rubric}
\end{figure*}
\clearpage

\clearpage
\begin{figure*}[p]
\centering

\begin{subfigure}[t]{0.5\textwidth}
\centering
\begin{tcolorbox}[
 title={Driver respect rubric prompt },
  boxrule=0.6pt,
  arc=1mm,
  left=1mm,right=1mm,top=1.5mm,bottom=1.0mm,
  width=\linewidth,
  % height=0.86\textheight,  % adjust to fit caption
  valign=top
]
\tiny

You are tasked with evaluating rationales explaining ratings of civilian respect in police traffic stops. Use the rubric below to guide your evaluation. The rubric describes three major categories of communication—Emotions, Professionalism, and Communication—along with Contextual Moderators that may shape them. These categories are not mutually exclusive or hierarchical. Instead, they capture overlapping aspects of civilian behavior and interaction that annotators may draw on when explaining their ratings.

Your role is to judge whether each of the respect dimensions are referenced in the provided rationale.

\textbf{1. Emotions and Attitudes}

Emotions and attitudes refer to how the civilian's language, tone, and demeanor convey warmth, empathy, or hostility. It highlights the affective dimension of respectful or disrespectful communication.

\textbf{Respectful Emotional Descriptives}

Warmth: The civilian conveys friendliness or politeness.\\
Example: ``The driver greeted the officer calmly and politely.''

Empathy/Understanding: The civilian acknowledges the officer's role or perspective.\\
Example: ``The driver said they understood the officer was just doing their job.''

Apology/Thanks: The civilian apologizes for inconvenience or expresses gratitude.\\
Example: ``They apologized for the mistake.''\\
Example: ``They thanked the officer for explaining the situation.''

\textbf{Disrespectful Emotional Descriptives}

Lack of Warmth: The civilian is cold, distant, or unfriendly.

Lack of Empathy/Understanding: The civilian doesn't show any understanding or concern.

Lack of Apology/Thanks: The civilian doesn't apologize or thank.

Offensiveness/Bias: The civilian uses insulting, hateful, or biased language.\\
Example: ``They insulted the officer's appearance.''

Unnecessary Escalation: The civilian amplifies tension rather than calming the situation.\\
Example: ``They started yelling aggressively even though the officer remained calm.''

Disrespect for Time: The civilian deliberately stalls or avoids moving the interaction forward.\\
Example: ``They refused to provide documents for a long time without explanation.''

\textbf{2. Professionalism}

Professionalism refers to composure, self-control, and respectful demeanor consistent with constructive interaction. It concerns whether the civilian behaves in a manner consistent with respectful standards.

\textbf{Respectful Professionalism}

Formal Address: The civilian uses titles such as ``Sir'' or ``Ma'am'' in a genuine, non-demeaning manner.\\
Example: ``The driver consistently addressed the officer as `sir'.''

Composure: The civilian stays calm under stress.\\
Example: ``They remained calm even when frustrated. They complained about the ticket but still complied and spoke respectfully.''

Polite Language: The civilian avoids profanity, sarcasm, or mocking tones.\\
Example: ``They spoke in clear and courteous language.''

Cooperation: The civilian complies with reasonable requests.\\
Example: ``They promptly provided their license and registration.''

\textbf{Disrespectful Unprofessionalism}

Aggressive Opening: The civilian begins with hostility.\\
Example: ``They started the interaction by shouting at the officer.''

Unprofessional Language: The civilian uses sarcasm, profanity, or mocking tones.\\
Example: ``They cursed at the officer repeatedly.''

Loss of Composure: The civilian becomes erratic or hostile.\\
Example: ``They slammed the car door and gestured aggressively.''

\textbf{3. Communication}

Communication refers to whether the civilian provides clarity, transparency, and space for dialogue.

\textbf{Respectful Communication}

Reason Given: The civilian provides their own account or explanation.\\
Example: ``They explained they were late to work.''

Honesty/Transparency: The civilian is forthcoming about their situation, background, or violations.\\
Example: ``He openly admitted being on probation and explained his circumstances calmly.''

Clarification Requests: The civilian asks questions respectfully.\\
Example: ``They asked politely what the next step was.''

Acknowledgment: The civilian shows they understood the officer's explanation.\\
Example: ``They said, `Okay, I understand'.''

\textbf{Disrespectful Communication}

Interrupts: The civilian cuts off or talks over the officer.\\
Example: ``They interrupted the officer repeatedly.''

Non-Responsive: The civilian refuses to answer or stonewalls.\\
Example: ``They ignored the officer's questions altogether.''

\textbf{4. Contextual Moderators}

Situational factors that shape how civilian communication and behavior are expressed. These are not inherently respectful or disrespectful but help explain why annotators may justify ratings differently.

Threaten Violence: The civilian makes verbal or physical threats.\\
Example: ``The driver threatened to harm the officer.''

Non-Visible Hands / Non-Compliance: The civilian hides hands or resists showing ID.\\
Example: ``They refused to show their hands when asked.''

Physical Resistance: The civilian resists stepping out of the vehicle.\\
Example: ``They locked the doors and refused to exit.''

Yelling: The civilian shouts over the officer.\\
Example: ``They screamed continuously, preventing dialogue.''

Extreme Interruptions: The civilian persistently cuts off the officer.\\
Example: ``They would not let the officer finish a sentence.''

Environmental Distractions: Noise or bystanders associated with the civilian impede communication.\\
Example: ``Their passengers shouted loudly throughout the stop.''

Limited Capacity: The civilian struggles to respond due to language barriers, intoxication, or confusion, but still attempts to cooperate.\\
Example: ``Although he was drunk and confused, he complied quickly and remained polite.''

\end{tcolorbox}
\caption{Part I.}
\end{subfigure}
\hfill
\begin{subfigure}[t]{0.49\textwidth}
\centering
\begin{tcolorbox}[
  boxrule=0.6pt,
  arc=1mm,
left=1.2mm,right=1.2mm,top=1.5mm,bottom=1.0mm,
  width=\linewidth,
  % height=0.86\textheight,
  valign=top
]
\tiny
\setlength{\parindent}{0pt}
\setlength{\parskip}{2pt}
\setlist[itemize]
{leftmargin=1.2em,topsep=1pt,itemsep=1pt,parsep=0pt}

\textbf{RATIONALE TO EVALUATE:}\\
\{rationale\}

\textbf{INSTRUCTIONS FOR OUTPUT}

After reading the rationale, evaluate it against the rubric above.\\
For each sub-dimension, answer the following question:\\
Is this aspect being referenced in the rationale, either explicitly or implicitly implied?\\
Respond with ``yes'' or ``no'' for each field.

When answering:
\begin{itemize}
  \item Treat respectful and disrespectful dimensions as separate and independent.
  \item Mark ``yes'' under a respectful field only if the rationale explicitly mentions the presence of that positive behavior.
  \item Mark ``yes'' under a disrespectful field only if the rationale explicitly mentions the absence or violation of that behavior.
  \item Do not mark both as ``yes'' unless the rationale clearly describes both respectful and disrespectful moments.
  \item If the rationale does not reference that aspect at all, mark ``no'' in both.
\end{itemize}

You must produce only valid JSON that conforms exactly to the schema below.

{\tiny
\begin{verbatim}
Schema :
{
  "emotional_respect": {
    "respectful": {
      "warmth": "",
      "empathy_understanding": "",
      "apology_thanks": ""
    },
    "disrespectful": {
      "lack_of_warmth": "",
      "lack_of_empathy_understanding": "",
      "lack_of_apology_thanks": "",
      "offensiveness_bias": "",
      "unnecessary_escalation": "",
      "disrespect_for_time": ""
    }
  },
  "professional_respect": {
    "respectful": {
      "formal_address": "",
      "composure": "",
      "polite_language": "",
      "cooperation": ""
    },
    "disrespectful": {
      "aggressive_opening": "",
      "unprofessional_language": "",
      "loss_of_composure": ""
    }
  },
  "communicative_respect": {
    "respectful": {
      "reason_given": "",
      "honesty_transparency": "",
      "clarification_requests": "",
      "acknowledgment": ""
    },
    "disrespectful": {
      "interrupts": "",
      "non_responsive": ""
    }
  },
  "contextual_moderators": {
    "threaten_violence": "",
    "non_visible_hands_non_compliance": "",
    "physical_resistance": "",
    "yelling": "",
    "extreme_interruptions": "",
    "environmental_distractions": "",
    "limited_capacity": ""
  }
}
\end{verbatim}
}
\end{tcolorbox}
\caption{Part II.}
\end{subfigure}
\caption{Rubric and instructions used by the LLM-as-a-judge to evaluate driver-respect rationales and output structured rubric activations.}
\label{fig:drvier_rubric}
\end{figure*}
\clearpage

\begin{figure*}
\centering
\begin{tcolorbox}[title={Officer system prompt for training and evaluation}, width=\textwidth, fontupper=\small]
You are a Black female evaluator, age:28. You or your family members are/were affiliated with law enforcement.

You will be given the transcript of a traffic stop in Los Angeles, and your task is to evaluate the primary officer's (labeled as Police:Primary in the transcript) respectfulness during this stop on a 1--5 scale:  
1 = Very Disrespectful,  
5 = Very Respectful.

First, provide a rating for the overall respectfulness of the {entity name} during the stop. Focus on the tone, language, and behavior displayed throughout the entire interaction. Consider the stop as a whole. A respectful interaction involves consistent professionalism, calm communication, and polite language, even in moments of disagreement or tension. Disrespectful behavior includes frequent interruptions, aggressive language, or dismissive actions. Reflect on how the interaction unfolded and whether the respect shown fluctuated or remained consistent throughout the stop. Consider both the officer’s and the civilian’s actions and demeanor, and provide an overall judgment based on the entire encounter.

When forming your judgment, consider four overlapping aspects of respect:  
- Emotional Respect: tone, warmth, empathy, or hostility  
- Professional Respect: fairness, procedural clarity, or abuse of authority  
- Communicative Respect: listening, interruptions, explanations, or clarity  
- Contextual Moderators: situational pressures or provocations that might affect judgment

Then, provide a rationale where you describe the specific situations and reasoning that led to your respect rating of the {entity name}. 
Describe concrete moments where the {entity name} demonstrated respectful or disrespectful behavior. Highlight key behaviors, tone shifts, or actions that influenced your overall judgment. 
Write the rationale naturally, as if explaining your judgment to another human annotator. Be specific and varied in focus—some rationales may emphasize tone, others fairness, others communication clarity. Avoid repeating the same phrasing across responses.

Ground your reasoning only in evidence from the transcript. Avoid generic statements.

Return output in exactly this format:  

Rating: <1--5>  

Rationale: <1--3 sentences>
\end{tcolorbox}

\caption{Example of prompt for training and evaluating our models on officer respect}
\label{fig:task_prompt}
\end{figure*}
% --- Preamble (add once) ---
% \usepackage{tcolorbox}
% \tcbuselibrary{skins,breakable}

\begin{figure*}[t]
\centering
\begin{tcolorbox}[
title={Augmenter module prompt},
, width=\textwidth, fontupper=\small
]
\textbf{System / Instruction.}  

You are an expert annotator trained to explain police officer respect ratings using the LAPD Respect Rubric. Your task is to paraphrase rationales that explain officer respect ratings in police traffic stops. You must rewrite them using different wording and sentence structure while preserving the same underlying meaning, overall rating, and rubric-dimension signals.

The paraphrased rationale should:
\begin{itemize}
    \item Reflect the same respect dimensions marked as present or absent in the provided dimension judgments;
    \item Preserve the original respect rating meaning;
    \item Use natural, fluent, human-like English;
    \item Vary wording, phrasing, and sentence structure from the original rationale.
\end{itemize}

\vspace{3pt}
\textbf{LAPD Respect Rubric}

The rubric defines three overlapping elements of respect used by annotators when explaining ratings:

\emph{Emotional Respect} captures warmth, empathy, apology, hostility, or unnecessary escalation;  
\emph{Professional Respect} concerns greetings, introductions, tone, and composure;  
\emph{Communicative Respect} reflects transparency, explanation, opportunities to speak, and clarity about next steps or termination of the encounter.

\vspace{3pt}
\textbf{User Input.}

\begin{itemize}
    \item Original Rating: \texttt{\{rating\}}
    \item Original Rationale: \texttt{\{rationale\}}
    \item Respect Dimension Judgments: \texttt{\{dimension\_json\}}
\end{itemize}

\textbf{Task.}  
Write three paraphrased rationales that preserve the same meaning, respect rating, and rubric-dimension alignment as the original rationale. Each paraphrase should differ in wording and structure while remaining semantically equivalent.

\textbf{Output Format.}  
Return \emph{only} a valid JSON object of the following form:
\begin{verbatim}
{
  "paraphrase_1": "...",
  "paraphrase_2": "...",
  "paraphrase_3": "..."
}
\end{verbatim}

Each paraphrase must be a single coherent paragraph. Do not introduce new behaviors or interpretations.
\end{tcolorbox}
\caption{Prompt used to generate rubric-preserving paraphrases of ground-truth rationales for preference pair construction.}
\label{fig:paraphrase_prompt}
\end{figure*}

\begin{figure*}[t]
\centering
\begin{subfigure}{\columnwidth}
\centering
\begin{tcolorbox}[title={Llama and Qwen prompt}, width=\linewidth, fontupper=\small]
You are going to get the rationales provided by annotators for the level of respect the believed was shown by a police officer to a civilian during a traffic stop in LA.
Your job is to predict the level of respect they shown based on their rationale.\\
Potential respect levels are: very disrespectful, disrespectful, neutral, respectful and very respectful. Respond with only one of these, and nothing else; no explanations, no clarifications, no questions.\\
The phrases very disrespectful, disrespectful, neutral, respectful and very respectful have been redacted from the rationales, and replaced with [MUTE].\\\\

Rationale: `This was a very routine interaction and the ofcr was a professional`\\
Assessment: very respectful
\end{tcolorbox}
\caption{Llama and Qwen prompt}
\label{fig:rr_pred_llama}
\end{subfigure}

\hfill

\begin{subfigure}{\columnwidth}
\centering
\begin{tcolorbox}[title={GPT-OSS prompt}, width=\linewidth, fontupper=\small]
<|start|>system<|message|>You are ChatGPT, a large language model trained by OpenAI.\\
Knowledge cutoff: 2024-06\\
Current date: 2025-09-24\\\\

Reasoning: low\\

\# Valid channels: analysis, commentary, final. Channel must be included for every message.<|end|><|start|>user<|message|>You are going to get the rationales provided by annotators for the level of respect the believed was shown by a police officer to a civilian during a traffic stop in LA.
Your job is to predict the level of respect they shown based on their rationale.\\
Potential respect levels are: very disrespectful, disrespectful, neutral, respectful and very respectful. Respond with only one of these, and nothing else.\\
The phrases very disrespectful, disrespectful, neutral, respectful and very respectful have been redacted from the rationales, and replaced with [MUTE].\\\\

Rationale: `The officer was professional and extremely helpful of the driver.`\\
<|end|><|start|>assistant<|channel|>final<|message|> Assessment: respectful
\end{tcolorbox}
\caption{GPT-OSS prompt}
\label{fig:rr_pred_gpt}
\end{subfigure}

\caption{Prompts used to evaluate rationale predictiveness of respect ratings across model families}
\label{fig:rr_pred_prompts}
\end{figure*}

\end{document}